\documentclass[conference]{IEEEtran}
\IEEEoverridecommandlockouts
\usepackage{cite}
\usepackage{amsmath,amssymb,amsfonts}
\usepackage{algorithmic}
\usepackage{graphicx}
\usepackage{subcaption}
\usepackage{textcomp}
\usepackage{xcolor}
\usepackage{hyperref}
\usepackage{multirow}
\usepackage{booktabs}
\usepackage[linesnumbered,ruled,vlined]{algorithm2e}

\def\BibTeX{{\rm B\kern-.05em{\sc i\kern-.025em b}\kern-.08em
    T\kern-.1667em\lower.7ex\hbox{E}\kern-.125emX}}
\begin{document}

\title{One Masked Model is All You Need for Sensor Fault Detection, Isolation and Accommodation*
\thanks{*This work was supported by GE Offshore Wind, part of GE Vernova.}
}

\author{
\IEEEauthorblockN{Yiwei Fu}
\IEEEauthorblockA{\textit{GE Vernova Advanced Research} \\
Niskayuna, NY, USA 12309 \\
\texttt{\href{mailto:yiwei.fu@ge.com}{yiwei.fu@ge.com}}}
\and
\IEEEauthorblockN{Weizhong Yan}
\IEEEauthorblockA{\textit{GE Vernova Advanced Research} \\
Niskayuna, NY, USA 12309 \\
\texttt{\href{mailto:yan@ge.com}{yan@ge.com}}}
}

\maketitle

\begin{abstract}
Accurate and reliable sensor measurements are critical for ensuring the safety and longevity of complex engineering systems such as wind turbines. In this paper, we propose a novel framework for sensor fault detection, isolation, and accommodation (FDIA) using masked models and self-supervised learning. Our proposed approach is a general time series modeling approach that can be applied to any neural network (NN) model capable of sequence modeling, and captures the complex spatio-temporal relationships among different sensors. During training, the proposed masked approach creates a random mask, which acts like a fault, for one or more sensors, making the training and inference task unified: finding the faulty sensors and correcting them. We validate our proposed technique on both a public dataset and a real-world dataset from GE offshore wind turbines, and demonstrate its effectiveness in detecting, diagnosing and correcting sensor faults. The masked model not only simplifies the overall FDIA pipeline, but also outperforms existing approaches. Our proposed technique has the potential to significantly improve the accuracy and reliability of sensor measurements in complex engineering systems in real-time, and could be applied to other types of sensors and engineering systems in the future. We believe that our proposed framework can contribute to the development of more efficient and effective FDIA techniques for a wide range of applications. 
\end{abstract}


\section{Introduction}~\label{sec:intro}
The advent of the Internet of Things (IoT) has ushered in a new era of interconnected devices, fostering the seamless exchange of information and enabling smart, data-driven decision-making across diverse domains. Sensors play a pivotal role in collecting and transmitting critical data for various IoT applications in many industries, including energy, aerospace, healthcare, and automotive. However, the reliability of these IoT applications hinges on the accurate functioning of sensors, which makes sensor fault detection, isolation, and accommodation (FDIA) paramount.  
Traditional FDIA approaches rely on expert knowledge and hand-crafted algorithms, which can be time-consuming and error-prone. In recent years, deep learning-based data-driven approaches have shown promising results in automating the FDIA process. However, most existing deep learning-based FDIA approaches require separate models for each task, which can be computationally expensive and difficult to integrate into existing systems. In addition, these approaches often require a specific neural network (NN) architecture, limiting their applicability to different types of sensors and systems.

To address these limitations, this paper presents a novel framework for sensor FDIA using self-supervised learning (SSL). Masked models and SSL have been popularized in recent years, particularly in the language domain, with the introduction of BERT (Bidirectional Encoder Representations from Transformers)~\cite{devlin_bert_2019}. BERT is a pre-trained masked language model that has achieved state-of-the-art performance on various natural language processing tasks, including question answering, sentiment analysis, and text classification. The masked language modeling approach used in BERT involves randomly masking some of the input tokens and training the model to predict the masked tokens based on the surrounding context. This approach has shown great success in capturing the complex relationships among different tokens in a sentence and has significantly improved the performance of language models.

Inspired by the success of masked models in the language domain, time series anomaly detection and forecasting tasks have also adopted masked models~\cite{fu2022mad, fu2022masked, fu2023masked}. The proposed framework in this paper uses a similar approach for sensor FDIA. During training, the proposed masked approach creates a random mask, which acts just like a fault, for one or more sensors. This key insight makes the training and inference task unified: finding the faulty sensors and correcting them. By using a masked model for sensor FDIA, the proposed framework can capture the complex spatiotemporal relationships among different sensors and perform all three tasks of sensor FDIA using a single NN model. This approach not only simplifies the overall FDIA pipeline but also outperforms existing approaches. By utilizing the masking technique, a single model can be used for implicit multi-task learning. As the mask changes, the task is also changing as FDIA on different sensors.

The proposed method was validated on a public dataset and a real-world turbine sensor FDIA use case. The masked model not only simplifies the overall FDIA pipeline, but also outperforms existing approaches.

The proposed framework has several advantages over existing approaches. First, it is a unified framework that performs all three tasks of sensor FDIA using a single NN model. This simplifies the overall FDIA pipeline and reduces computational costs. Second, the proposed approach is not limited to a specific NN architecture, making it a general approach applicable to all deep learning time series models (model-agnostic). Third, the unified training and inference pipeline improves the accuracy and efficiency of fault detection and correction. In contrast, most existing approaches only train the model on normal data, making the inference step with faulty sensor data essentially an out-of-distribution problem.

The main contribution of our paper is on introducing masked modeling (SSL) technique to a new domain or application - sensor FDIA. The proposed framework has the potential to transform the way industries approach sensor fault detection and correction.

The rest of the paper is organized as follows: Section II provides a literature review of existing FDIA approaches, highlighting their limitations and the need for a unified framework. Section III describes the proposed framework in detail, including the masked approach during training. Section IV presents the experimental results, including the validation on several public datasets and a real-world turbine sensor FDIA use case. Finally, Section V concludes the paper and discusses future work.

\section{Related Work}~\label{sec:review}
\subsection{Sensor Fault Detection, Isolation and Accommodation}~\label{sec:review:fdia} 
Existing FDIA methods can be broadly classified into three categories: model-based, data-driven, and hybrid approaches. Model-based methods rely on physical models of the system and sensors to detect faults, but they require accurate modeling and may not be effective for all types of faults. Data-driven methods use statistical techniques and machine learning algorithms to detect faults, but they may not be effective for fault isolation and accommodation. Hybrid methods combine model-based and data-driven approaches to improve effectiveness and efficiency. 


Over the years, many different data-driven sensor fault detection methods have been proposed, including statistical analysis-based, e.g., principal component analysis (PCA)\cite{Berbache_2019},\cite{Li_2019} and independent component analysis (ICA) \cite{KIM_2013},  expert system-based, e.g., rules and fuzzy rules \cite{Chanak_2016}, and machine learning-based, e.g., support vector machine (SVM) \cite{Zidi_2018} and various types of neural networks \cite{Darvishi_2020},\cite{Hussain_2015}, including deep neural networks \cite{Jana_2022}, \cite{cth2_12366}.  Among these diverse neural networks, auto-associative neural networks (AANN) might be the one most widely used for sensor validation. For example, in \cite{ELNOUR_2020} AANN was used for sensor validation and fault diagnosis for building HVAC systems. It used both control signals and operating condition variables in addition to sensor measurements as the inputs to the AANN, but reconstructed 6 sensor measurements only. In \cite{Vanini_2014}, a bank of AANNs were used for performing fault detection and isolation for both sensor and system faults.

Most of existing data-driven FDIA methods have shown reasonably good performance in terms of sensor fault detection; however, for sensor fault localization/isolation as well as accommodation, most of these methods require using many models. Jana et al.~\cite{jana2022cnn} use a Convolutional Neural Network (CNN) classifier for fault detection, and multiple Convolutional Autoencoder (CAE) networks for contstruction of each fault type. However, this method is limited to a pre-defined number of fault types, and may struggle with a new fault type. Darvishi et al.~\cite{darvishi2020sensor} developed a complicated architecture for sensor FDIA with multiple estimators and predictors, a residual calculator and a multi-layer perceptron (MLP)-based classifier. In contrast to these existing methods, our proposed method uses a single model for all three tasks of sensor FDIA, making it more efficient and effective, and is not limited to a specifically-designed architecture for a limited number of faults.

\subsection{Masked Time Series Models}~\label{sec:review:mask} Masked models have been popularized in recent years, particularly in the language domain, with the introduction of BERT (Bidirectional Encoder Representations from Transformers)~\cite{devlin_bert_2019}. BERT is a pre-trained masked language model that has achieved state-of-the-art performance on various natural language processing tasks, including question answering, sentiment analysis, and text classification. The masked language modeling approach used in BERT involves randomly masking some of the input tokens and training the model to predict the masked tokens based on the surrounding context. This approach has shown great success in capturing the complex relationships among different tokens in a sentence and has significantly improved the performance of language models.

In addition to the language domain, masked models have also shown promising results in the time series domain, especially in anomaly detection. For example, Fu et al. proposed a general masked anomaly detection task for time series data~\cite{fu2022mad}. The proposed self-supervised learning task outperformes traditional next step prediction task for time series anomaly detection using the same base model. Xu et al.~\cite{xu2023masked} proposed a masked graph neural network for anomaly detection in time series data. It learns the structure of sensor networks with a graph, and then uses masked tempral learning for handling the time series.

Other masked time series models have been applied to forecasting problems. Fu et al. also proposed a masking technique for general multivariate multi-step time series forecasting~\cite{fu2022masked}. The proposed approach is flexible and can incorporate known future information in the forecast, making it better than traditional sample-based forecasting or recursive forecasting. Tang et al.~\cite{tang2022mtsmae} proposed a masked autoencoder for time series forecasting with a patch embedding method. Another use of masked time series model is for pre-training. Zha et al.~\cite{zha2022time} proposed a masked autoencoder with extrapolator for self-supervised time series generation. The authors claim it performs well in downstream tasks such as time series classification, prediction and imputation. Dong et al.~\cite{dong2023simmtm} proposes to recover masked points by weighted aggregration of neighbors outside the manifold, and learns local structure of the manifold. The pre-trained model is then used for both forecasting and classification. Li et al.~\cite{li2023ti} trained another masked autoencoder for time series reconstruction, and used it for downstream forecasting and classification tasks.

Inspired by the success of masked models in the language and time series domains, in this paper we adopt the masked modeling idea to a novel application domain - sensor FDIA. In addition to simply detecting a fault, our proposed method can also isolate the faulty sensor and accommodate it by replacing the faulty sensor reading with the predicted value from the model.

\section{Proposed Method}~\label{sec:method}
\begin{figure}[t]
    \centering
    \begin{subfigure}[b]{0.95\linewidth}
        \includegraphics[width=\textwidth]{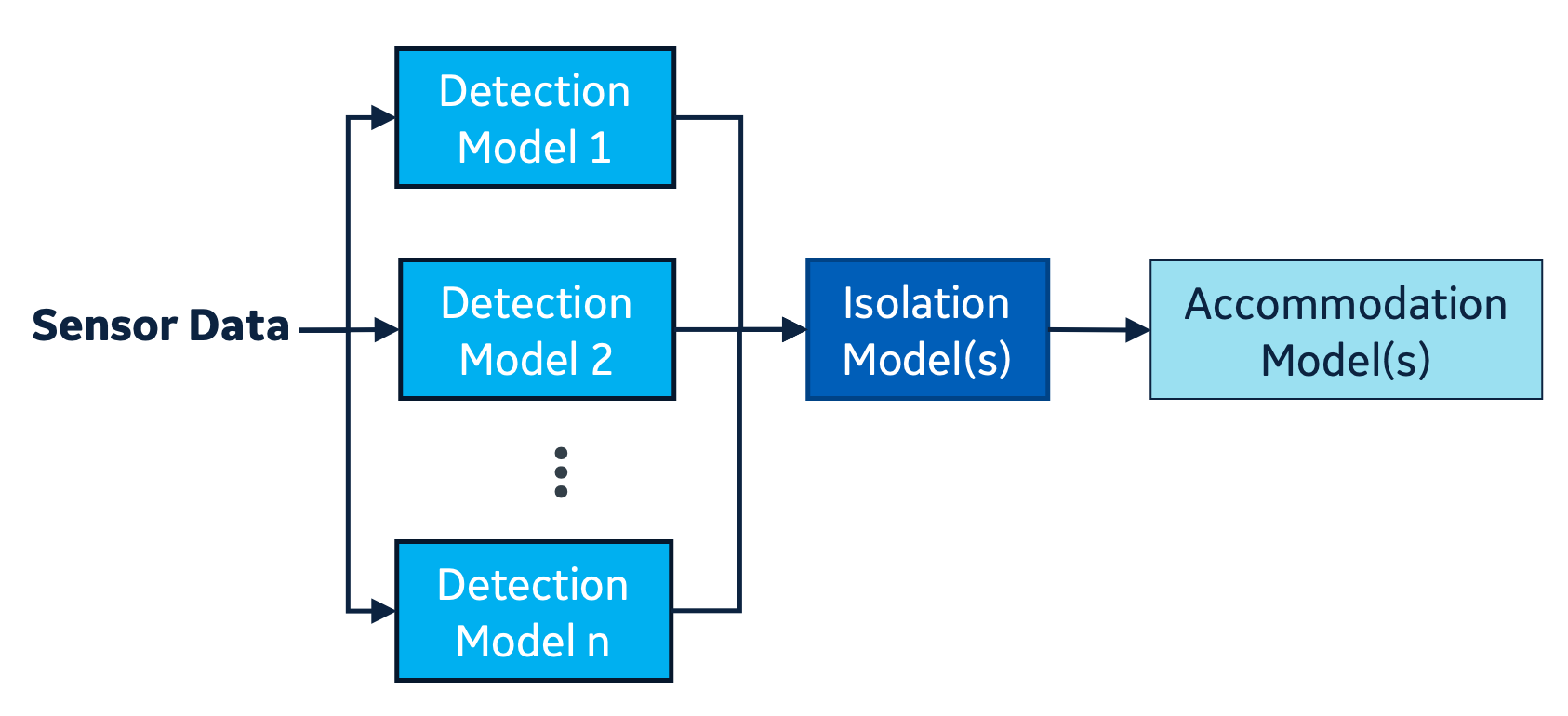}
        \caption{Existing sensor FDIA schemes typically require separate models for detection, isolation, and accommodation.}
        \label{fig:formulation:prior_art}
    \end{subfigure}
    \hfill
    \begin{subfigure}[b]{0.95\linewidth}
        \includegraphics[width=\textwidth]{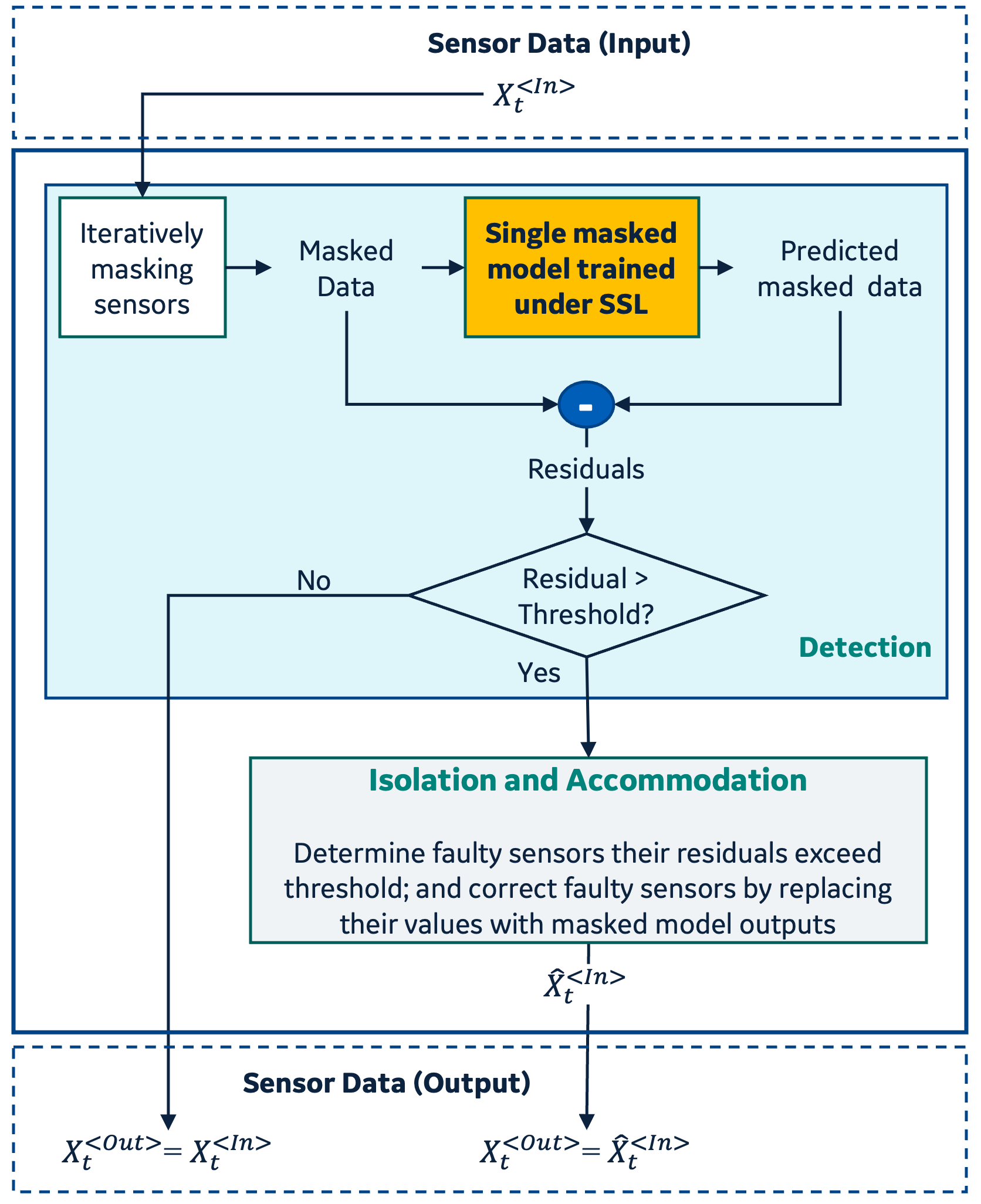}
        \caption{Our proposed masked model formulation uses a single model to complete all three tasks of sensor FDIA.}
        \label{fig:formulation:invention}
    \end{subfigure}
    \caption{Comparison of existing sensor FDIA techniques with our proposed method.}
    \label{fig:formulation}
\end{figure}

Traditional sensor FDIA frameworks typically require separate models for each of the three stages of the process: detection, isolation, and accommodation, as shown in Figure~\ref{fig:formulation:prior_art}. The detection model is responsible for identifying when a fault or failure has occurred, while the isolation model is responsible for determining which sensor(s) are faulty or failing. Finally, the accommodation model is responsible for taking corrective action to ensure that the system continues to operate safely and effectively. While this approach has been relatively effective in many applications, it can be complex and time-consuming to develop and maintain separate models for each stage of the process. Instead, we propose a more integrated FDIA framework that can streamline the process and improve overall system performance. Our approach involves using a single masked model to model the sensor data, as shown in Figure~\ref{fig:formulation:invention}. Once trained, this model can be used to complete all three tasks of sensor FDIA. In the following subsections, we delve into the details of our approach.

\subsection{Modeling Sensor Data with Masked Models}~\label{sec:method:training}
We denote $\mathbf{x_t}\in \mathbb{R}^n$ as the sample from the $n$-channel sensor time series at time $t$, with the $i$-th dimension at time $t$ denoted as $x^i_t$ (i.e., $\mathbf{x_t}=[x_t^1, x_t^2,...,x_t^n]$). Under self-supervised learning (SSL), there are several different formulations to model the normal sensor data, including \textit{auto-regressive} and \textit{auto-associative} formulations, as illustrated in Figure~\ref{fig:training}, in addition to recently developed masked modeling. Our proposed method, shown in Figure~\ref{fig:training:subfig1} and detailed in Algorithm~\ref{algo:train}, randomly masks one or more channels (up to $0.2n$) for all time steps during training, with the task of reconstructing the masked channel(s) using the rest. Note that the loss is calculated only on the masked channels and not on the known channels, and the masks are randomly chosen at each iteration. 

For the standard auto-regressive formulation in Figure~\ref{fig:training:subfig2}, the task is to generate all sensor data at time $t$ using previous information. In the auto-associative (e.g., autoencoder) formulation as shown in Figure~\ref{fig:training:subfig3}, the task is to reconstruct all the sensor data at time $t$ with itself by using a bottleneck layer. In practice, for time series models in our masked approach and the auto-regressive approach, a window of length $T$ is often used instead of all historical samples prior to time $t$. This window length can be adjusted to different applications and datasets.

\begin{figure}
  \centering
    \begin{subfigure}[b]{1.0\linewidth}
      \begin{equation*}
      \begin{aligned}
        \left[
        \begin{aligned}
          \rule{1em}{1em}\\[-0.5em]
          x^2\\[-0.5em]
          \rule{1em}{1em}\\[-1em]
          \vdots\\[-0.5em]
          \rule{1em}{1em}
        \end{aligned}
        \right]_{t, t-1, t-2, ...} &= f\left(
        \left[
        \begin{aligned}
          x^1\\[-0.5em]
          \rule{1em}{1em}\\[-0.5em]
          x^3\\[-1em]
          \vdots\\[-0.5em]
          x^n
        \end{aligned}
        \right]_{t, t-1, t-2, ...}
        \right),
        \\
        \left[
        \begin{aligned}
          x^1\\[-0.5em]
          \rule{1em}{1em}\\[-0.5em]
          \rule{1em}{1em}\\[-1em]
          \vdots\\[-0.5em]
          x^n
        \end{aligned}
        \right]_{t, t-1, t-2, ...} &= f\left(
        \left[
        \begin{aligned}
          \rule{1em}{1em}\\[-0.5em]
          x^2\\[-0.5em]
          x^3\\[-1em]
          \vdots\\[-0.5em]
          \rule{1em}{1em}
        \end{aligned}
        \right]_{t, t-1, t-2, ...}
        \right),
      \cdots  
      \end{aligned}
      \end{equation*}
      \vspace{-8pt}
      \caption{Our proposed masked formulation. The black squares indicate masked data during training. In each iteration, we randomly mask one or more channels and use the remaining channels to compute them. The loss is calculated only on the masked channel(s).}
      \label{fig:training:subfig1}
    \end{subfigure}
    \hfill
    \begin{subfigure}[b]{0.55\linewidth}
      \begin{equation*}
        \left[
        \begin{aligned}
          x^1\\[-1em]
          \vdots\\[-0.5em]
          x^n
        \end{aligned}
        \right]_t = f\left(
        \left[
        \begin{aligned}
          x^1\\[-1em]
          \vdots\\[-0.5em]
          x^n
        \end{aligned}
        \right]_{t-1, t-2, ...}
        \right)
      \end{equation*}
      \vspace{-8pt}
      \caption{Auto-reegressive.}
      \label{fig:training:subfig2}
    \end{subfigure}
    \hfill
    \vline
    \hfill
    \begin{subfigure}[b]{0.43\linewidth}
      \begin{equation*}
        \left[
        \begin{aligned}
          x^1\\[-1em]
          \vdots\\[-0.5em]
          x^n
        \end{aligned}
        \right]_t = f\left(
        \left[
        \begin{aligned}
          x^1\\[-1em]
          \vdots\\[-0.5em]
          x^n
        \end{aligned}
        \right]_{t}
        \right)
      \end{equation*}
      \vspace{-8pt}
      \caption{Auto-associative.}
      \label{fig:training:subfig3}
    \end{subfigure}
  \caption{Training: Comparison of different sensor data modeling formulations. Superscripts indicate channel, subscripts indicate time step, and $f$ indicates a mapping or a model. Our proposed masked method (a) achieves multi-task learning by randomly masking one or more channels at each iteration, in contrast to existing techniques (b) Auto-regressive and (c) Auto-associative.}
  \label{fig:training}
\end{figure}

\begin{algorithm}[!ht]
	\KwIn{Time series model $f_{\mathbf{\theta}}$ with trainable parameters $\mathbf{\theta}$, maximum history length $T$, loss function $\ell$}
	\KwData{Time series dataset $S=\{\mathbf{x_t}\}$, where $\mathbf{t}$ represents time $t$, and $\mathbf{x_t}\in \mathbb{R}^n$}
	Preprocess dataset with a sliding window of length $(T+1)$ to $\{\mathbf{x_{t-T}},...,\mathbf{x_{t-1}},\mathbf{x_{t}}\}$ sequences\;
	Randomly initialize model parameters $\mathbf{\theta}$\;
	\While{not at end of training epochs}{
		\While{not at the end of all mini-batches}{
			Randomly choose an integer number of channels $c_m$ for this batch, $0<c_m\leq 0.2n$\;
			\For{each sequence in the mini-batch}{
				Randomly choose $c_m$ channels to mask;
			}
			Feed masked sequences to model $f_{\mathbf{\theta}}$, generate estimations $\hat{\mathbf{x}}^{c_m}$\ with unmasked channels;
			Calculate loss $\sum_{^{c_m}}\ell(\mathbf{x}^{c_m},\hat{\mathbf{x}}^{c_m})$\;
			Backpropagation, update model parameters $\mathbf{\theta}$\;
		}
	}
	\KwOut{Trained model $f_{\mathbf{\theta}}$}
	\caption{Masked Sensor FDIA Training}
	\label{algo:train}
\end{algorithm}

Our proposed masked formulation is superior to existing approaches for sensor FDIA, as it effectively deals with faulty sensors during training. By masking a channel, the model is forced to handle faulty sensors and accomplish the accommodation task. This approach uses one model and different masks to learn the comprehensive relationships among different sensors, leading to the same training and inference task.

In contrast, auto-regressive or auto-associative formulations suffer from the out-of-distribution problem during inference, as they are only modeled on normal data. When the input contains faulty sensor data, masked models are better equipped to handle them, as they are designed to do so during training. Additionally, autoencoders require a bottleneck layer to limit their capacity and avoid learning a trivial identity mapping. However, this leads to information loss and worse performance. In our approach, we avoid the need for a bottleneck layer when it is not necessary, resulting in better performance and more efficient training. Autoencoders also calculate the loss on all sensors, which include the normal sensors. This is not ideal, as the model should not be penalized for not reconstructing normal sensors. In contrast, masked models only calculate the loss on the masked sensors, which are the ones that need to be reconstructed.

Masked models are able to utilize training data more effectively. This is because the same sequence can be masked in different ways, creating more scenarios for different faults. In practice, for time series models, we often mask by replacing the original sensor values with random values within the range for that channel, rather than using a fixed value like 0, since 0 has a physical meaning in many sensors~\cite{fu2022mad}.

During the masked model training phase, we typically mask one channel and sometimes multiple channels up to $0.2n$, assuming that most faults occur at a single sensor. As the number of sensors masked increases, the training task becomes increasingly difficult. This is not a limitation of our formulation, but more of a limitation of available information which also applies to all existing sensor FDIA approaches. If more than half of the sensors are faulty, neither regression nor autoencoder formulations can handle the situation. In this case, it is more sensible to raise an alarm after the localization step and conduct an investigation, rather than attempting to accommodate all the faulty sensors.

If some sensors are not relevant to the FDIA pipeline but could still be useful in the models, they can be included in the input without being masked. This means that in Figure~\ref{fig:training}, these sensors would be added to the right side of the equation but not masked, and not being added to the left side.

\subsection{Online Inference for Sensor FDIA}~\label{sec:method:inference}
Once a masked model is trained, it can be used for online inference, as shown in Algorithm~\ref{algo:inference}. 
The process involves sequentially masking each sensor and calculating the predicted values for that masked sensor with all other sensors. The model output is then compared to the sensor reading, and if the residual is greater than a threshold, the sensor is considered faulty. In this case, the faulty sensor reading is replaced with the model output, which is expected to represent the normal sensor reading when there is no fault on that channel.

\begin{algorithm}[!ht]
	\KwIn{Trained time series model $f_{\mathbf{\theta}}$ from Algorithm~\ref{algo:train}, threshold $th^i$ for each channel $i$}
	\KwData{A sequence of length $(T+1)$ sensor data $\{\mathbf{x_{t-T}},...,\mathbf{x_{t-1}},\mathbf{x_{t}}\}$, each with $n$ channels}
  \For{each channel $i$ in range $1$ to $n$}{
    $\hat{\mathbf{x}}^i = f_\theta(\mathbf{x}^c)$, where $c=\{k \mid k \neq i, 1\leq k \leq n\}$\;
    (Detection, Isolation) \If{$\frac{\lvert\hat{\mathbf{x}}^i-\mathbf{x}^i\rvert}{T+1}>th^i$}{
      (Accomodation) $\mathbf{x}^i = \hat{\mathbf{x}}^i$\;
    }
    }
	\KwOut{Sensor data $\mathbf{x}$ with faulty sensors replaced by model predictions}
	\caption{Masked Sensor FDIA Online Inference}
	\label{algo:inference}
\end{algorithm}

It is worth noting that although we are demonstrating channel-wise FDIA, our method is also capable of determining when a fault begins by identifying the first time step where the residual exceeds the threshold. Despite the need to run the inference $n$ times for $n$ sensors, our proposed method achieves all three tasks of fault detection, isolation, and accommodation with a single model and $n$ forward passes. As a result, the inference time only grows linearly with the number of sensors. In contrast, traditional sensor FDIA formulations, as shown in Figure~\ref{fig:formulation}, require multiple detection, isolation, and accommodation models, making inference more expensive.

Moreover, if multiple sensors are found to be faulty, we can mask all of them and recalculate the predictions to address the issue. However, as discussed in Section~\ref{sec:method:training}, if too many sensors are faulty during inference, it is advisable to raise an alarm after the localization step in a real-world system. This is because a large number of faulty sensors could indicate a serious problem that requires immediate attention, and the issue becomes less of a machine learning problem at that point.

In cases where a large number of sensors are found to be faulty, it is crucial to conduct a thorough investigation to identify the root cause of the problem and take appropriate measures to address it. This may involve replacing faulty sensors, repairing damaged equipment, or taking other corrective actions. Attempting to infer the data of a large number of failed sensors is not a practical solution, as it can lead to inaccurate results and potentially dangerous situations. By raising an alarm and conducting a detailed investigation, it is possible to prevent serious accidents or equipment failures, ensuring the safety and reliability of the system.

\section{Experiments and Results}~\label{sec:exp}
In this section, we evaluate our proposed method for sensor FDIA from two perspectives using two different base models. First, we present the results of our proposed method and compare it against other formulations specified in Figure~\ref{fig:training} using a publicly available dataset. The purpose of this comparison is to validate the effectiveness of our proposed method for detecting sensor faults. We then provide an example of applying our method in a real-world complex engineering system to demonstrate its practical use in creating a more robust and reliable system.

\subsection{Sensor Fault Detection with HAI Dataset}~\label{sec:exp:public}
We first demonstrate the effectiveness of our proposed method for \textit{sensor fault detection} using the HAI dataset~\cite{10.5555/3485754.3485755}. The dataset~\footnote{HAI dataset can be downloaded at \url{https://github.com/icsdataset/hai}.} contains time-series data collected from a realistic industrial control system (ICS) testbed augmented with a Hardware-In-the-Loop (HIL) simulator that emulates steam-turbine power generation and pumped-storage hydropower generation. The dataset consists of several CSV files, each of which contains the recorded values of a set of SCADA (Supervisory Control and Data Acquisition) data points at a given time. This study is conducted based on the $20.07$ version of HAI dataset, which has a total of $n=59$ process measurements.

The HAI testbed consists of four industrial processes: Boiler Process (P1), Turbine Process (P2), Water-treatment Process (P3), and HIL Simulation (P4). To demonstrate the effectiveness of our method, we selected four potential sensors, namely P1\_PIT01 (Heat-exchanger outlet pressure), P1\_PIT02 (Water supply pressure of the heating water pump), P1\_TIT01 (Heat-exchanger outlet temperature), and P1\_TIT02 (Temperature of the heating water pump). These sensors were chosen because they are critical for monitoring the performance of the processes and are susceptible to sensor faults. During training and inference, only these four sensor measurements were masked, while the other 55 sensor measurements were used to provide additional information to the model, which can help improve the accuracy of the predictions. 

We utilized the first normal operation data file, `train1.csv', comprising 309,600 samples, and divided it into training and validation sets at an 80\%-20\% ratio. For the second file, `train2.csv', containing 241,200 samples, we introduced two minor faults to validate our method's effectiveness. We added a 0.2 bar bias to sensor P1\_PIT01 (range: $[0.5678, 2.4332]$ bar) and a 1-degree Celsius bias to sensor P1\_TIT01 (range: $[34.7046, 36.7340]$ degrees). We compared these faults against the original normal operating conditions. All time series data were scaled to a $[0, 1]$ range before training, using a time window size of 20 for both training and inference.

For fair comparison, we used an encoder-only Transformer\cite{vaswani_attention_2017} model as the base machine learning model for both our proposed masked formulation and regression. The Transformer model had a model dimension of 64, feed-forward dimension of 256, number of heads of 8, and number of layers of 2. For the autoencoder formulation, we used a 3-layer autoencoder with 64, 32, and 2 dimensions, with the 2-dimensional bottleneck layer representing the latent space. Mean Square Error (MSE) loss was used for all models with an Adam optimizer\cite{kingma_adam_2015} and a learning rate of 0.001. All models were trained for 200 epochs. We then used the trained model to detect the faults and compared the results with the ground truth.

\begin{figure}[htbp]
    \centering
    \begin{subfigure}[b]{0.49\linewidth}
      \includegraphics[width=\textwidth]{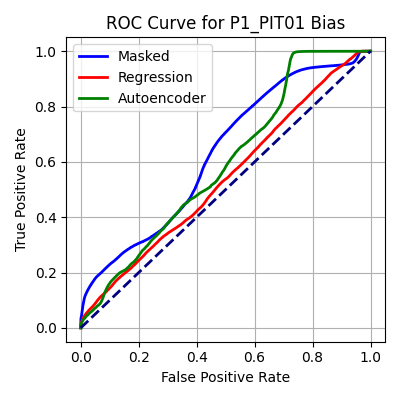}
      \label{fig:roc_p1_pit01}
    \end{subfigure}
    \begin{subfigure}[b]{0.49\linewidth}
      \includegraphics[width=\textwidth]{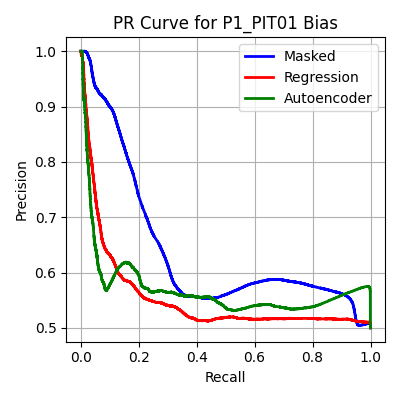}
      \label{fig:pr_p1_pit01}
    \end{subfigure}
    
    \begin{subfigure}[b]{0.49\linewidth}
      \includegraphics[width=\textwidth]{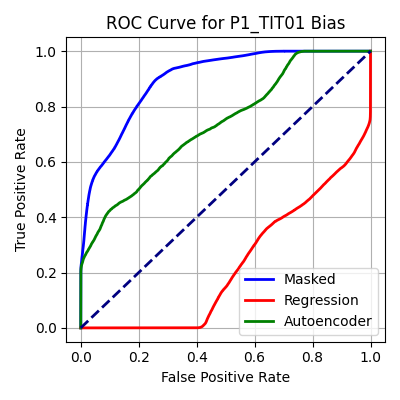}
      \label{fig:roc_p1_tit01}
    \end{subfigure}
    \begin{subfigure}[b]{0.49\linewidth}
      \includegraphics[width=\textwidth]{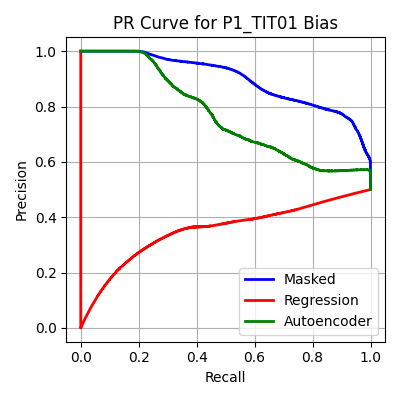}
      \label{fig:pr_p1_tit01}
    \end{subfigure}
    \caption{ROC and PR curves for sensor bias in P1\_PIT01 and P1\_TIT01.}
    \label{fig:hai_results}
  \end{figure}

After training the model with normal data, we used the trained model to detect the bias sensor faults against the original normal data. Both the area under curve for ROC curve (ROC AUC) and the area under curve for precision-recall curve (AUPRC) were used to evaluate the performance of the models. The ROC curve is a plot of the true positive rate (TPR) against the false positive rate (FPR) at various threshold settings, while the precision-recall curve is a plot of the precision (TPR) against the recall (FPR) at various threshold settings. The ROC AUC and AUPRC values range from 0 to 1, with a higher value indicating better performance. The results are shown in Table~\ref{table:hai_results}, and corresponding plots of the ROC and precision-recall curves are shown in Figure~\ref{fig:hai_results}.

\begin{table}[ht]
	\centering
	\caption{HAI dataset sensor fault detection results. ROC AUC and AUPRC are reported, with the best in bold for each fault.}
		\begin{tabular}{c|c|c|c}
			\toprule
			Sensor Bias & Method & ROC AUC & AUPRC\\
			\midrule
			\multirow{3}{*}{P1\_PIT01} & Masked & \textbf{0.6300} & \textbf{0.6410} \\ 
			& Regression & 0.5408 & 0.5533 \\
			& Autoencoder & 0.5932 & 0.5714 \\
			\midrule
			\multirow{3}{*}{P1\_TIT01} & Masked & \textbf{0.8996} & \textbf{0.8981} \\
			& Regression & 0.2264 & 0.3521 \\ 
			& Autoencoder & 0.7342 & 0.7669 \\ 
			\bottomrule
		\end{tabular}
	\vspace{0pt}
	\label{table:hai_results}
\end{table}

Our proposed masked formulation outperforms the regression formulation by a large margin for small sensor faults, using the same base Transformer model and hyperparameters. For the P1\_PIT01 sensor bias, our method achieves ROC AUC and AUPRC values of 0.6300 and 0.6410, respectively, compared to 0.5408 and 0.5533 for the regression formulation. For the P1\_TIT01 sensor bias, our method achieves ROC AUC and AUPRC values of 0.8996 and 0.8981, respectively, compared to 0.2264 and 0.3521 for the regression formulation. Our method is able to detect faults even with small biases, as it is trained to predict sensor values with masked sensor values. The poor results from regression for the P1\_TIT01 sensor bias may be due to the small size of the bias, which the regression formulation is not able to detect. The autoencoder formulation also performs relatively well. However, it is not as effective as our proposed method. This demonstrates the effectiveness of our proposed method for sensor fault detection.

\subsection{Sensor FDIA inside Wind Turbine Blade Load Estimation Pipeline}~\label{sec:exp:ge}
\begin{figure}[!thb]
    \centering
    \begin{subfigure}[b]{0.86\linewidth}
        \includegraphics[width=\textwidth]{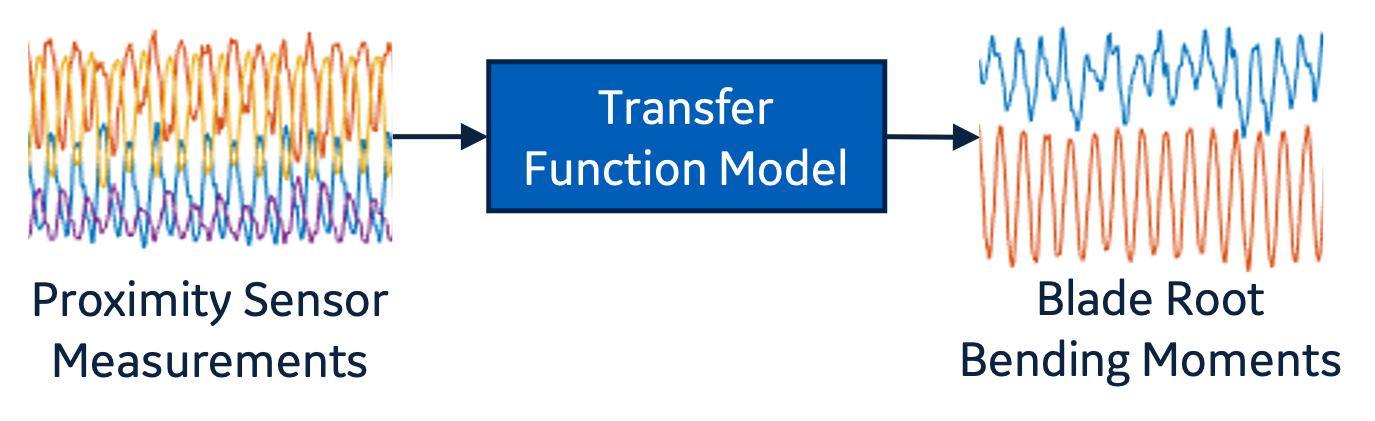}
        \caption{Existing blade root load estimation pipeline.}
        \label{fig:exp:ge_old}
    \end{subfigure}
    \hfill
    \begin{subfigure}[b]{0.9\linewidth}
        \includegraphics[width=\textwidth]{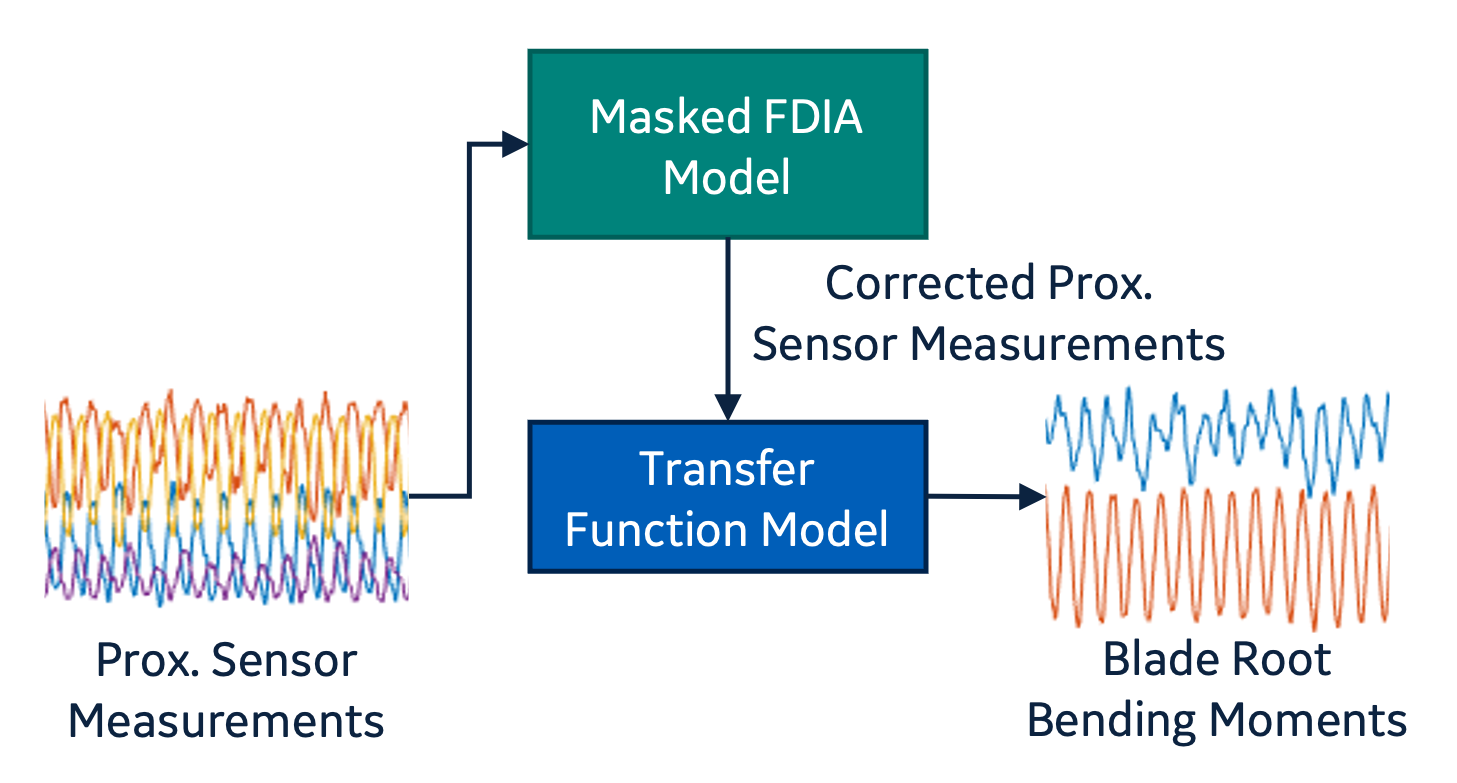}
        \caption{Improved robust blade root load estimation pipeline with our proposed sensor FDIA technique.}
        \label{fig:exp:ge_new}
    \end{subfigure}
    \caption{Example of how our proposed sensor FDIA technique can be integrated into an existing engineering system to improve its robustness and reliability.}
    \label{fig:ge}
\end{figure}

A wind turbine is a complex engineering system that relies on accurate and reliable sensor measurements for turbine control and for achieving turbine's optimal performance. In this study, we present a real-world application case study where the proposed FDIA technique is applied to sensor measurements of GE offshore wind turbines, ensuring accuracy and reliability of the sensor measurements by performing FDIA in real-time. Specifically, we focus on the FDIA of proximity sensors, which are used for estimating blade root bending moments. Blade bending moments are the critical measurements for effective blade load control of wind turbines. The accuracy and reliability of these measured moments are crucial for ensuring the safety and longevity of wind turbines, as well as maximizing their energy output. As directly measuring blade bending moments is costly and unreliable, we at GE have developed a novel, cost-effective solution for obtaining the blade bending moments reliably. It uses proximity sensor measurements to infer blade bending moments via a transfer function (TF) model, as shown in Figure~\ref{fig:exp:ge_old}.  While the TF model can be either a physics-based model or a data-driven model, for this study, we use a data-drive TF model and keep it fixed for all of experiments so that we can focus on the FDIA of the proximity sensor measurements. Our proposed sensor FDIA technique is integrated in this pipeline to detect, diagnose, and correct proximity sensor faults in real-time, as shown in Figure~\ref{fig:exp:ge_new}, and thus enabling the pipeline to be more robust and reliable in obtaining blade bending moments.

More specifically, each blade has four proximity sensors located at different angles, resulting in a total of 12 input and 12 output sensors for the Masked FDIA Model shown in Figure~\ref{fig:exp:ge_new}. The TF model takes the 12 proximity sensor measurements, along with five additional sensor measurements (including pitch angles for the three blades, the rotor azimuth angle, and the rotor speed) as inputs, and outputs the six estimated blade root bending moments (two for each blade). We use a normal operation dataset provided by GE Offshore Wind, which is split into training, validation, and test sets. The training set consists of 1,610,031 samples, the validation set consists of 223,778 samples, and the testing set consists of 462,305 samples. The raw data is sampled at 25 Hz, and we downsampled it to 1 Hz for masked sensor FDIA.

For the FDIA model in this application, we used a simple 2-layer LSTM~\cite{hochreiter_long_1997} model with 50 hidden units each. Tims series sequence length is 10 seconds, and other hyperparameters are similar to those in Section~\ref{sec:exp:public}. For the sensor fault cases, we introduced 4 different faults:
\begin{itemize}
    \item \textbf{Bias}: A proximity sensor on blade 1 is biased by 1 standard deviation of the sensor values.
    \item \textbf{Drift}: A proximity sensor on blade 1 drifts from 0 to 2 standard deviation of the sensor values.
    \item \textbf{Noise}: A proximity sensor on blade 1 is added a white noise of 2 standard deviation of the sensor values.
    \item \textbf{Bias Two Sensors}: Two proximity sensors on blade 1 and blade 3 are both biased by 1 standard deviation of their respective sensor values.
\end{itemize}

\begin{figure}[htbp]
    \centering
    \begin{subfigure}[b]{0.49\linewidth}
      \includegraphics[width=\textwidth]{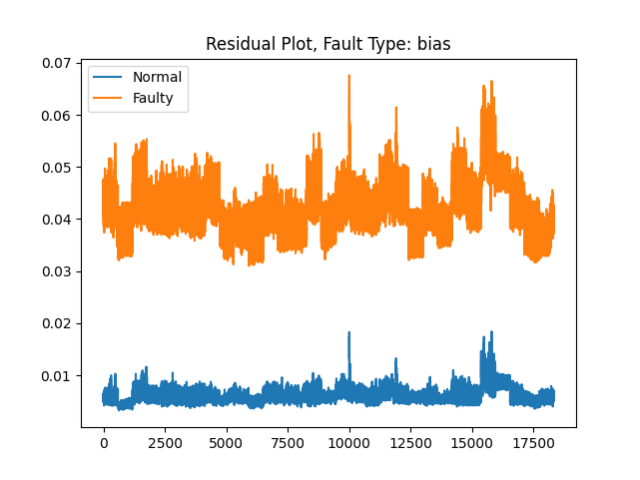}
      \caption{Fault type: Bias.}
      \label{fig:turbine_results:bias}
    \end{subfigure}
    \hfill
    \begin{subfigure}[b]{0.49\linewidth}
      \includegraphics[width=\textwidth]{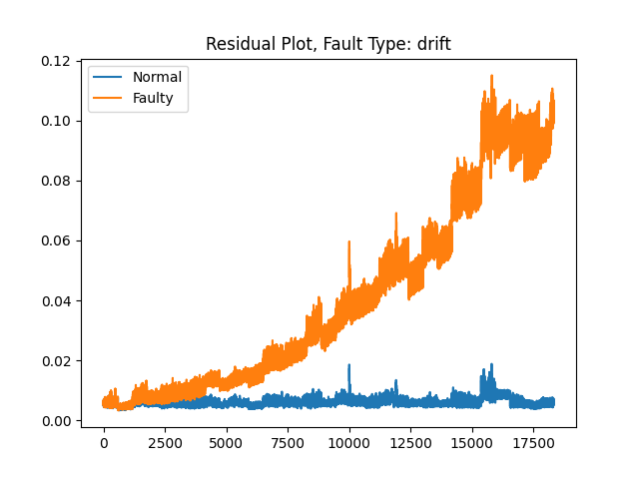}
      \caption{Fault type: Drift.}
      \label{fig:turbine_results:drift}
    \end{subfigure}
    
    \begin{subfigure}[b]{0.49\linewidth}
      \includegraphics[width=\textwidth]{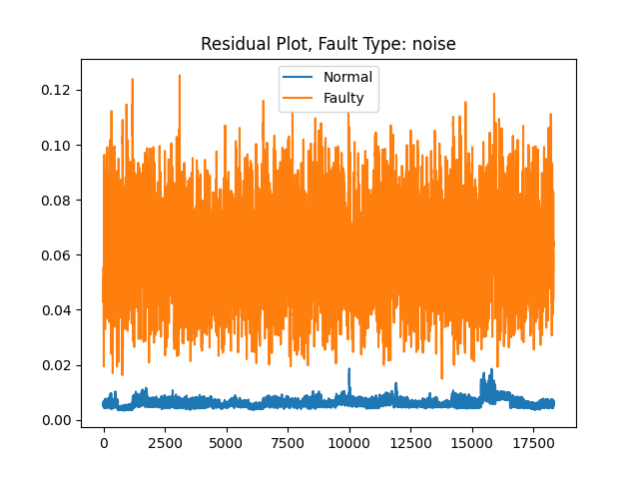}
      \caption{Fault type: Noise.}
      \label{fig:turbine_results:noise}
    \end{subfigure}
    \hfill
    \begin{subfigure}[b]{0.49\linewidth}
      \includegraphics[width=\textwidth]{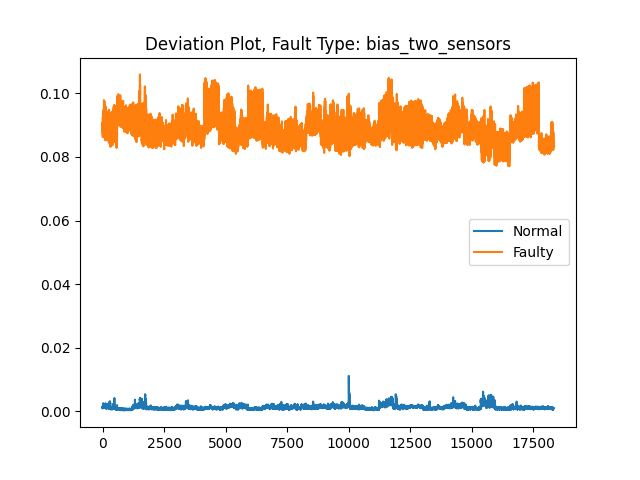}
      \caption{Fault type: Bias Two Sensors.}
      \label{fig:turbine_results:bias_2_sensors}
    \end{subfigure}
    \caption{Residual plots for the 4 different sensor fault cases versus the normal test set.}
    \label{fig:turbine_results}
  \end{figure}

After we train a masked sensor FDIA model, we test it on the normal test set versus the four different sensor faults. The results are shown in Figure~\ref{fig:turbine_results}. As shown in the residual plots, our proposed masked sensor FDIA model is able to detect all four sensor faults. If we look at the residual for each sensor, we can then isolate the faulty sensor. Afterwards, we can accommodate the faulty sensor by replacing its value with the predicted value from the FDIA model.

To further illustrate the impact of proximity sensor FDIA in the blade load estimation pipeline, we compare the performance of the existing pipeline (Figure~\ref{fig:exp:ge_old}) versus the improved pipeline with our proposed sensor FDIA technique (Figure~\ref{fig:exp:ge_new}). As a baseline, we put the normal test set through the existing pipeline and obtain the blade bending moment estimate errors. We then test the TF model with faulty bias data, both with (as in Figure~\ref{fig:exp:ge_new}) and without sensor FDIA (as in Figure~\ref{fig:exp:ge_old}). The results are shown in Table~\ref{table:ge_sensor_fault_accomodation}.

\begin{table*}[th]
	\centering
	\caption{Comparing performance of the existing pipeline versus the improved pipeline with our proposed sensor FDIA technique when sensor fault happens. We report the MSE between the predicted load vs true load for each blade and each bending moment type after the same Transfer Function (TF) model, in unit kNm. Lower is better.}
		\begin{tabular}{c|c|c|c|c}
			\toprule
			\multirow{2}{*}{Blade} & Bending & \multicolumn{3}{c}{Same TF Model Load Prediction MSE} \\ 
            & Moment & No Fault (Baseline) & Faulty Data w/o FDIA & Faulty Data w/ FDIA \\
			\midrule
			\multirow{2}{*}{\#1} & Flap & 276.29 & 2676.64 (+868.78\%) & 295.93 (+7.11\%) \\ 
			& Edge & 373.31 & 1656.74 (+348.80\%) & 394.04 (+5.55\%) \\
			\midrule
			\multirow{2}{*}{\#2} & Flap & 315.54 & 3537.59 (+1021.12\%) & 347.23 (+10.04\%) \\
			& Edge & 262.59 & 3148.68 (+1099.09\%) & 284.60 (+8.38\%) \\
            \midrule
			\multirow{2}{*}{\#3} & Flap & 326.11 & 2078.76 (+537.44\%) & 345.99 (+6.10\%) \\
			& Edge & 320.09 & 1645.86 (+414.19\%) & 384.28 (+20.05\%) \\ 
			\bottomrule
		\end{tabular}
	\vspace{0pt}
	\label{table:ge_sensor_fault_accomodation}
\end{table*}

Our proposed sensor FDIA technique significantly improves the robustness and reliability of the blade load estimation pipeline, as demonstrated by the reduction in blade bending moment estimate errors compared to the existing pipeline. In contrast to the existing pipeline, our technique is able to detect and correct sensor faults in real-time, and can even accommodate faulty sensor data to a reasonable degree. Our masked FDIA model not only corrects faults, but also raises alarms for specific faulty sensors, which can be used to trigger maintenance and repair.

The results of our evaluation show that the existing pipeline is not robust to sensor faults, and the blade bending moment estimate errors are significantly increased as the sensor faults occur, which can lead to catastrophic failures in the wind turbine. In some cases, when there is a bias fault in one of the proximity sensors, the blade bending moment estimate errors are increased by more than 10 times, which is unacceptable. In contrast, with our proposed sensor FDIA technique, the blade bending moment estimate errors are significantly reduced, and the performance is comparable to the baseline (no fault) case. Overall, our proposed sensor FDIA technique has the potential to greatly improve the performance and safety of wind turbines. By reducing downtime and generating more profits, our technique can also have a significant economic impact.

Furthermore, for this sensor FDIA application to be implemented in a real-time wind turbine monitoring system, the inference time of the FDIA model is also an important factor. We tested the model inference time of our proposed sensor FDIA model on an office desktop computer with Nvidia Quadro P2000 GPU. The average inference time of 600 runs through the dataset for our masked FDIA model is 9.2 ms. The inference time of the TF model is 1.6 ms. Although the FDIA model adds 5.75x inference time to the pipeline, the total (10.8ms) is still much less than the 40 ms (25 Hz) requirement for real-time monitoring. Comparing this to the huge reduction in blade bending moment estimate errors shown in Table~\ref{table:ge_sensor_fault_accomodation}, the trade-off is totally worth it. This also shows that our proposed sensor FDIA technique can be implemented in a real-time wind turbine monitoring system.

\section{Conclusion}~\label{sec:conclusion}
In this paper, we proposed a novel machine learning-based Fault Detection, Identification, and Analysis (FDIA) framework for sensor measurements in complex engineering systems using masked model. Our proposed technique is designed to detect, diagnose, and correct sensor faults in real-time, which is critical for ensuring the accuracy and reliability of sensor measurements and preventing costly downtime and maintenance. It is a general framework that can be applied to a wide range of sensor types and engineering systems. Any deep learning-based time series models would benefit from switching to our proposed masked formulation from the traditional, commonly-used regression formulation. 

Our proposed machine learning-based FDIA technique was validated on both a public dataset and a real-world dataset from GE offshore wind turbines. The technique effectively detected and diagnosed sensor faults in real-time, and corrected these faults to improve the reliability of sensor measurements. We focused on the FDIA of proximity sensors for estimating blade loads in GE offshore wind turbines, which is critical for ensuring the safety and longevity of wind turbines, as well as maximizing their energy output. Our evaluation of the technique on real-world sensor data demonstrated its effectiveness in detecting and diagnosing sensor faults in real-time, and its potential for preventing costly downtime and maintenance.

In conclusion, our proposed machine learning-based FDIA technique has the potential to significantly improve the accuracy and reliability of sensor measurements in a wide range of complex engineering systems. Future work could explore the application of our technique to other types of sensors and engineering systems, as well as the integration of our technique with other machine learning-based approaches for predictive maintenance and fault diagnosis.

\section*{Acknowledgment}
The authors would like to thank GE Offshore Wind for providing the data used in this study.

\bibliographystyle{IEEEtran}
\bibliography{sensor_fdia}

\end{document}